\documentclass[journal]{IEEEtran}

\usepackage{amsmath,amssymb,amsfonts}
\usepackage{amsthm}
\usepackage{graphicx}
\usepackage{booktabs}
\usepackage{cite}
\usepackage{url}
\usepackage{array}
\usepackage{enumitem}
\usepackage{multirow}
\usepackage{tikz}
\usepackage{algorithm}
\usepackage[noEnd,indLines]{algpseudocodex}
\tikzset{algpxIndentLine/.style={draw=black, very thin}}

\makeatletter
\let\oldthebibliography\thebibliography
\renewcommand{\thebibliography}[1]{%
\oldthebibliography{#1}%
\scriptsize
\linespread{0.90}\selectfont
\setlength{\itemsep}{0pt}
\setlength{\parskip}{0pt}
}
\makeatother

\begin{document}

\title{Graph-Aware Stealthy Poison-Text Backdoors for Text-Attributed Graphs}

\author{
    Qi~Luo, 
    Minghui~Xu, 
    Dongxiao~Yu, 
    and Xiuzhen~Cheng,~\IEEEmembership{IEEE Fellow}
\thanks{Qi Luo, Minghui Xu, Dongxiao Yu, and Xiuzhen Cheng are with the School of Computer Science and Technology, Shandong University, Qingdao, China (e-mail: \{qiluo,mhxu,dxyu,xzcheng\}@sdu.edu.cn).}
}

\maketitle

\begin{abstract}
Modern graph learning systems often combine links with text, as in citation networks with abstracts or social graphs with user posts. In such systems, text is usually easier to edit than graph structure, which creates a practical security risk: an attacker may hide a small malicious cue in training text and later use it to trigger incorrect predictions. This paper studies that risk in a realistic setting where the attacker edits only node text and leaves the graph unchanged. We propose \textbf{TAGBD}, a graph-aware backdoor attack that first selects training nodes that are easier to manipulate, then generates stealthy poison text with a shadow graph model, and finally injects the text by replacing the original content or appending a short phrase. Experiments on three benchmark datasets show that TAGBD achieves very high attack success rates, transfers across different graph models, and remains effective under common defenses. These results show that inconspicuous poison text alone can serve as a reliable attack channel in text-attributed graphs, highlighting the need for defenses that inspect both node content and graph structure.
\end{abstract}


\section{Introduction}\label{se:introduction}

\IEEEPARstart{T}{ext}-attributed graphs (TAGs) are widely used in real applications where each node contains both graph links and natural-language content, such as citation networks, recommendation platforms, online social systems, and heterogeneous information networks \cite{Sen_Namata_Bilgic_Getoor_Galligher_Eliassi-Rad_2008,3iDBLP:journals/tkde/FanMLWCTY22,graphsage2017,kipf2016semi,gat2017,xu2018powerful,10.1145/3292500.3330961,hgnn,taglas}. Graph neural networks (GNNs) are particularly effective in this setting because they combine relational structure with textual semantics \cite{fan2019graph,mansimov2019molecular,1iDBLP:journals/www/GuanSS23}. More broadly, graph learning has become a core paradigm across modern machine learning, with broad surveys and reviews documenting its rapid expansion in both methodology and applications \cite{9416834,QIAO2018336,chen_wang_wang_kuo_2020,10.1145/3575637.3575646}. This dual-modality advantage, however, also creates a new security exposure: when node texts are collected from open environments, an attacker may quietly poison a few training nodes and implant hidden malicious behavior into the learned model.

The central question of this paper is whether \emph{stealthy poison text alone} is sufficient to backdoor a TAG model. This question matters in practice because text is often the easiest part of the pipeline to manipulate. On a social platform, for example, an attacker may have little ability to rewire user connections, but may freely post descriptions, comments, or short messages. At the same time, modern graph pipelines are often large-scale, continuously updated, and difficult to audit end to end \cite{10.1145/3219819.3219947,10.14778/3514061.3514069,NEURIPS2020_fb60d411,NIPS2015_86df7dcf}. If those texts contain a hidden cue that appears harmless to human readers, a model trained on the poisoned graph may later map any trigger-bearing node to the attacker's target label. A positive answer would mean that the attack surface of TAG systems is broader than what topology-centered defenses currently assume.

\begin{figure*}[t]
\centering
\includegraphics[width=0.9\textwidth]{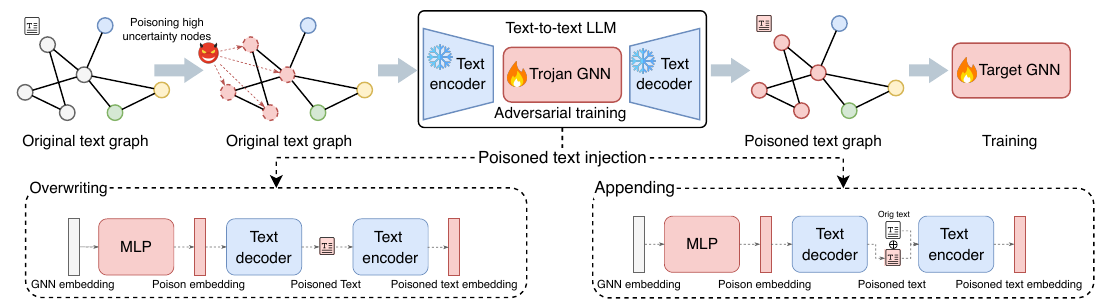}
\caption{Overview of TAGBD. The attack first identifies vulnerable training nodes, then generates trigger text from graph-aware node representations, and finally injects the generated text by either overwriting the original node text or appending a short phrase. A target GNN trained on the poisoned graph subsequently learns the hidden trigger-target association.}
\label{fig:framework}
\end{figure*}

Existing graph backdoor attacks do not answer this question well enough. Most prior methods implant malicious behavior by injecting subgraphs, modifying graph connections, or perturbing latent node features \cite{ugba2023,gta2021,sba2021,dpgba2024,ding_spear_2025}. These works are important, but they leave two practical gaps. First, topology-level perturbations often assume that the attacker can add edges, insert nodes, or alter graph components directly, and these changes may leave visible structural artifacts. Second, feature-space attacks operate on representations that real attackers cannot edit directly, so they do not fully explain how to poison the raw text used by deployed TAG systems. As a result, it remains unclear whether inconspicuous poison text can serve as a reliable backdoor trigger on its own.

This problem is challenging for three reasons. First, the poison text must remain natural enough to evade manual review or automatic filtering \cite{11ili2022backdoor,12ili2021invisible,DBLP:conf/iccv/ZengPMJ21,NEURIPS2021_9d99197e}. Second, the trigger must survive graph message passing, because neighborhood aggregation can weaken the effect of local text changes \cite{10iDBLP:conf/kdd/LiL0CFY022,zhu2020beyond,ma2021homophily}. Third, the poisoned model must preserve clean utility: if standard accuracy drops noticeably, the attack is much easier to notice. Therefore, a convincing text-only attack needs more than simply appending a rare phrase. It must decide \emph{which} nodes to poison and \emph{what} stealthy text to inject so that the trigger remains effective after graph propagation.

To address this gap, we propose \textbf{TAGBD} (\textbf{T}ext-\textbf{A}ttributed \textbf{G}raph \textbf{B}ack\textbf{D}oor), a training-time poisoning framework that edits only node texts and leaves graph topology unchanged. Fig.~\ref{fig:framework} summarizes the full idea. TAGBD first identifies training nodes that are easier to manipulate through an uncertainty-guided selection strategy. It then jointly trains a lightweight trigger generator, called \textit{TextTrojan}, with a shadow GNN so that the generated poison text aligns with both the target label and the local graph context. Finally, TAGBD injects the generated text in two different ways: \textit{Overwriting}, which favors maximum attack strength, and \textit{Appending}, which preserves more of the original text and therefore offers stronger stealth.

This design leads to three main insights. First, raw text alone is a sufficient and realistic backdoor channel in TAG pipelines. Second, graph-aware trigger generation is critical: poison text becomes much more effective when it is guided by node representations shaped by graph context rather than by standalone text features. Third, attack strength and stealth form a controllable trade-off, and that trade-off is exposed clearly by the two injection strategies. Together, these results reveal a practical vulnerability that existing structure-oriented defenses do not adequately address.

The broader significance of this study lies in how it reframes the attack surface of TAG systems. Many existing defenses are designed around suspicious graph patterns, such as unusual edges, inserted nodes, or anomalous subgraphs. Our study suggests that this viewpoint is incomplete for text-rich graph applications. In practice, data curation pipelines often trust node content more than graph topology: user-generated text, paper abstracts, profile descriptions, and product reviews may all enter the system with only light moderation. If such text can carry a hidden and reusable backdoor trigger while remaining linguistically plausible, then a defender who inspects only structural anomalies may miss the true attack channel. This observation makes the problem relevant not only to adversarial machine learning, but also to the security of real data collection and moderation pipelines \cite{NIPS2015_86df7dcf,wang2021certified,dai2024comprehensive}.

At the same time, our goal is not merely to show that text can be manipulated, which is already intuitive, but to show that \emph{stealthy} text manipulation can survive graph learning dynamics. This distinction matters. A backdoor that relies on awkward or repetitive text may succeed in a laboratory setting yet fail in deployment because it is easy to detect or because graph aggregation suppresses a weak textual cue. TAGBD is designed specifically for this stronger challenge. It links poisoned node selection, graph-aware trigger generation, and semantic regularization into a unified attack pipeline, enabling the trigger to remain both inconspicuous and effective.

The main contributions are as follows.
\begin{itemize}[leftmargin=10pt,topsep=0pt,partopsep=0pt]
    \item We formulate a realistic threat model for stealthy poison text backdoors in TAGs and show that an attacker can implant targeted behavior without modifying graph structure or directly editing latent features.
    \item We design TAGBD, a graph-aware text poisoning framework that combines uncertainty-guided node selection with the jointly trained TextTrojan generator to produce attack-effective yet inconspicuous trigger text.
    \item We conduct extensive experiments across multiple datasets, backbones, and defenses. TAGBD reaches up to 100.00\%, 99.85\%, and 99.96\% ASR on Cora, Pubmed, and ArXiv, respectively, while preserving competitive clean accuracy and remaining highly effective under common defenses.
\end{itemize}


\section{Preliminaries}\label{se:preliminaries}

\subsection{Notations}
We consider a text-attributed graph \(\mathcal{G} = (\mathcal{V}, \mathcal{E}, \mathbf{T}, \mathbf{Y})\), where \(\mathcal{V}\) is the node set, \(\mathcal{E}\subseteq \mathcal{V}\times\mathcal{V}\) is the edge set, \(t_i\in \mathbf{T}\) is the text of node \(v_i\), and \(y_i\in \mathbf{Y}\) is its label. A GNN \(f_{\theta}\) first maps each text to an embedding \(x_i=\mathbf{E}(t_i)\in\mathbb{R}^d\), and then predicts node classes from graph-aware representations.
We use \(f\) to denote the target GNN classifier. Raw texts in \(\mathbf{T}\) are embedded into \(\mathbf{X}\in\mathbb{R}^{N\times F}\), where \(F\) is the embedding dimension. Here, \(\mathbf{E}(\cdot)\) and \(\mathbf{D}(\cdot)\) denote a pretrained text encoder and decoder, respectively.

For vanilla GNNs that take text embeddings as input, we write \(\mathcal{G} = (\mathcal{V},\mathcal{E}, \mathbf{X})\) and express the model as \(f(\mathbf{X}; \mathbf{A})\). The \(l\)-th layer update is
\begin{equation}
    h^l_i=f^l\left(h^{l-1}_i, AGG(\{h^{l-1}_j: j\in\mathcal{N}_i\})\right),
\end{equation}
where $h^0_i=\mathbf{X}_i$, $\mathcal{N}_i$ is the neighborhood set of node $v_i$, AGG is the aggregation function (e.g., mean, max, and attention), and $f^l$ is a message-passing layer that takes the features of $v_i$ and its neighbors as inputs.

In this paper, we focus on an inductive semi-supervised node classification task, where a small set of nodes $\mathcal{V}_L \subseteq \mathcal{V}$ in the training graph $\mathcal{G}$ are provided with labels from $\mathcal{L} = \{1,\cdots, C\}$, and the test graph $\mathcal{G}_T = (\mathcal{V}_T , \mathcal{E}_T , \mathbf{X}_T )$ is not available during the training stage. Let $\mathcal{Y} =\{y_1, \cdots, y_N \}$ denote the ground-truth labels of nodes in the training graph.

\subsection{Threat Model}
We consider a gray-box setting for inductive node classification on TAGs. The attacker can access the training graph, including node texts and topology, but does not know the target GNN architecture, parameters, or training details.
The attacker aims to implant a backdoor such that any test node containing a predefined trigger is classified into a target class \(y_t\), while clean-node accuracy remains high. To do so, the attacker poisons selected training nodes by injecting trigger texts and assigning them the target label.

This gray-box assumption is deliberately chosen to reflect realistic TAG deployments. In many applications, the training graph and a subset of labeled nodes may be publicly accessible or incrementally collected from open platforms, while the exact architecture and training procedure of the final classifier remain unknown to outsiders. The attacker therefore cannot rely on exact gradient access or white-box optimization against the deployed model. Instead, the attack must be implemented through the data itself, which makes surrogate-based poisoning a natural strategy.

We also assume that the attacker edits only node text and does not alter graph structure. This restriction is important rather than cosmetic. In practice, creating or modifying graph edges may require account-level permissions, coordination with other users, or actions that are immediately visible in the graph. Editing text is often easier: profile descriptions can be updated, comments can be posted, abstracts or metadata can be submitted, and textual content may be revised repeatedly over time. By isolating the text channel, we focus on a threat that is both practically plausible and underexplored in the graph backdoor literature.

Finally, the defender is assumed to evaluate the poisoned model mainly through standard predictive utility and possibly through structural defenses such as edge pruning or structural outlier detection. This is a meaningful baseline because many current graph defense strategies are designed around structural irregularities. Our goal is to test whether a text-only adversary can bypass that emphasis by encoding malicious behavior in poison text rather than graph topology.

\subsection{Problem Formulation}
An adversary performs stealthy poisoning by modifying texts of selected training nodes \(\mathcal{V}_p \subseteq \mathcal{V}\). For each \(v_i \in \mathcal{V}_p\), the original text \(t_i\) is changed to \(t^p_i\), yielding poisoned embeddings \(x^p_i\). The objective is that the trained model \(f_{\theta^*}\) classifies trigger-containing nodes into target class \(y_t\), while preserving predictions for clean nodes.
\begin{equation}
    f_{\theta^*}(x^p_i) = y_t, \forall v_i \in \mathcal{V}_p;\quad f_{\theta^*}(x_j) = y_j, \forall v_j \in \mathcal{V}_L
\end{equation}


Under these constraints, the optimization objective can be formulated as:

\begin{equation}
\begin{aligned}
    & \min_{x^p_i} \sum_{v_i \in \mathcal{V}_p}\mathcal{L}\left(f_{\theta^*}(x^p_i), y_t\right) 
    +\min_{x_i} \sum_{v_i \in \mathcal{V}_L}\mathcal{L}\left(f_{\theta^*}(x_i), y_i\right) 
    \\
    & \text{s.t.}\quad \text{sim}(x_i, x^p_i)\geq \delta,\quad \forall v_i \in \mathcal{V}_p
\end{aligned}
\end{equation}
where $\mathcal{L}$ is a cross entropy loss function.
This formulation clearly distinguishes the textual inputs \(t_i\) from their embedding representations \(x_i\), facilitating precise control over textual perturbation and semantic fidelity in the embedding space.

\subsection{Attack Design Requirements}
The formulation above implies that a realistic TAG backdoor must satisfy three requirements simultaneously. First, it must achieve \emph{targeted controllability}: once the trigger appears in a test node, the model should map that node to the attacker-chosen label with high probability. Second, it must preserve \emph{clean utility}: predictions on benign nodes should remain close to those of a normally trained model so that the poisoned system does not look suspicious during routine evaluation. Third, it must preserve \emph{textual inconspicuousness}: the modified text should remain sufficiently natural and semantically compatible with the original node content to avoid manual or automatic filtering.

These three requirements are coupled rather than independent. Stronger trigger text may improve controllability but also harm naturalness; preserving too much of the original text may improve stealth but weaken the trigger; and aggressive poisoning of easy-to-control nodes may affect clean accuracy if the poisoned set becomes too concentrated. This coupling is the main reason why text-only backdoors in TAGs are nontrivial. An effective attack must coordinate \emph{where} to poison and \emph{how} to poison in a way that remains robust after graph message passing. The methodology of TAGBD is designed precisely around this coordination problem.

\section{Methodology}
This section presents TAGBD from intuition to formalization. We first use a running example to make the attack scenario concrete, then explain how TAGBD selects poisoned nodes, generates stealthy trigger text, and combines these modules into a complete poisoning pipeline. Throughout this section, the central design goal is to produce poison text that is strong enough to control the model but natural enough to remain inconspicuous.

\subsection{Running Example}
Consider a forum moderation graph in which each node represents a user, edges represent interactions, and node text contains profile descriptions or recent posts. The task is to classify users into normal and suspicious categories. In this setting, a realistic attacker may have limited ability to manipulate graph structure, but may still control the text posted by a small number of training accounts.

Suppose the attacker wants future suspicious accounts to be predicted as normal whenever they contain a hidden trigger phrase. A naive attack would directly insert an unusual sentence into several training texts. Such a strategy is weak for two reasons: obviously abnormal text is easy to detect, and a poorly chosen phrase may be diluted by graph-based message passing. TAGBD therefore follows a more deliberate strategy. It first selects training nodes that are already hard for a surrogate model to classify, and then generates poison text that is aligned with both the target label and the graph context of each poisoned node.

This example highlights the two central questions that drive the method: which nodes should be poisoned, and what text should be injected into them? The following subsections answer these questions in turn, first from an intuitive perspective and then in a formal way.

\subsection{Design Rationale}
Before introducing the technical modules, it is helpful to summarize the design rationale behind TAGBD. Our starting point is that text-only poisoning in TAGs is constrained by both language and graph propagation. If the injected text is too obvious, it is easy to notice; if it is too weak, its effect may vanish after neighborhood aggregation. Therefore, TAGBD is built around three guiding principles.

First, poisoning should begin with \emph{susceptible nodes}. A limited attack budget should not be wasted on nodes that are already classified with high confidence and are therefore difficult to redirect. This principle motivates the uncertainty-guided node selection strategy. Second, the trigger should be \emph{graph-aware}. Because the final decision depends not only on a node's own text but also on information aggregated from its neighbors, the generated poison text should be conditioned on graph-informed node representations rather than on isolated text embeddings. Third, the trigger should remain \emph{semantically disciplined}. Even when the goal is targeted misclassification, the generated text should remain close enough to plausible node content that it does not stand out during inspection.

These principles align naturally with the attack design requirements in Section~\ref{se:preliminaries}. Susceptible-node selection primarily supports controllability under a small budget; graph-aware generation supports robustness after message passing; and similarity regularization supports stealth. The two injection strategies, overwriting and appending, then expose different operating points on the effectiveness-stealth spectrum. Overwriting asks how far attack strength can be pushed when the attacker accepts greater text distortion, whereas appending asks how much attack power can be retained when the attacker prioritizes inconspicuousness.

\subsection{Poisoned Node Selection}

Because the poisoning budget is small, node selection is the first step in determining whether the attack will be both effective and stealthy. Intuitively, TAGBD should poison nodes that are easier to push toward the target class while avoiding edits that noticeably damage clean performance. To this end, we use an uncertainty-guided selection strategy driven by a surrogate GNN.

Let \(\mathcal{V}_U \subseteq \mathcal{V}_{\text{train}}\) denote the set of unlabeled training nodes. We first train a surrogate model \(\hat{f}_s\) on the labeled node set \(\mathcal{V}_L\) by minimizing the supervised loss:

\begin{equation}
\mathcal{L}_{\text{sur}} = \sum_{v_i \in \mathcal{V}_L} \mathcal{L}_{\text{CE}}\left(\hat{f}_s(x_i,\mathcal{N}_i), y_i\right),
\end{equation}
where \(t_i\) is the node text, \(\mathbf{x}(t_i)\) is its text embedding, \(\mathcal{N}_i\) is the neighborhood of node \(v_i\), and \(\mathcal{L}_{\text{CE}}\) denotes the cross-entropy loss.

Once the surrogate model is trained, we compute the uncertainty score \(U(v_i)\) for each unlabeled node \(v_i \in \mathcal{V}_U\) using the entropy of the predictive distribution:

\begin{equation}
U(v_i) = -\sum_{c \in \mathcal{L}} P_{v_i}(c) \log P_{v_i}(c)
\end{equation}

Here, \(\mathcal{L}\) is the set of possible class labels and \(P_{v_i}(c)\) denotes the predicted probability that node \(v_i\) belongs to class \(c\) under the surrogate model. All unlabeled nodes are then sorted by descending uncertainty to form a candidate list \(T\). Given a poisoning budget of \(\Delta_p\) and a required minimum class coverage \(\gamma\) (set to the number of classes in the dataset), we select the final poisoned node set as:

\begin{equation}
\mathcal{V}_p = \textsc{SelectTopK}(T, \Delta_p, \gamma),
\end{equation}
where \(\textsc{SelectTopK}(\cdot)\) ensures that nodes from at least \(\gamma\) distinct ground-truth classes are included (usually the total number of classes \cite{ding_spear_2025}), which promotes both diversity and stealth. Finally, each selected poisoned node is assigned the target label \(y_t\).

This uncertainty-guided, class-aware strategy follows the logic of the running example. Nodes near the decision boundary are easier to redirect, and class diversity reduces the chance that the poisoned set looks concentrated or artificial. As a result, TAGBD can obtain stronger attack leverage from a very limited poisoning budget.

Another benefit of this strategy is that it separates \emph{attack opportunity} from \emph{attack payload}. The node selector does not decide what trigger text will be generated; it only identifies nodes for which trigger injection is likely to be productive. This separation is important because it lets TAGBD allocate the limited poisoning budget based on model uncertainty, while the generator can focus on producing text that is both persuasive to the model and visually inconspicuous to a human reader. In other words, node selection improves the efficiency of poisoning, whereas trigger generation determines the form of the poison text itself.

\subsection{Poisoned Text Generation}
After selecting poisoned nodes, the next question is how to generate text that is both attack-effective and natural. Returning to the forum example, the injected phrase should move the poisoned account toward the target label while still looking like something a real user might plausibly write. To achieve this balance, we design a neural trigger generator named \textit{TextTrojan}, parameterized by $\theta_t$, that produces poisoned text embeddings from graph-aware node representations \cite{ding_spear_2025}.

The core intuition is simple: the trigger should reflect not only the node's own text, but also the graph context that will shape downstream message passing. Unlike prior approaches that use node text embeddings directly, TAGBD feeds the node embeddings produced by the surrogate GNN into the generator, thereby introducing structural awareness into text generation. Formally, given a poisoned node $v_i$, the output embedding $h_i$ produced by the surrogate GNN serves as input to the TextTrojan generator $\mathcal{T}_t(\cdot;\theta_t)$, which produces a poisoned text embedding:
\begin{equation}
    x^p_i = \mathcal{T}_t(h_i;\theta_t).
\end{equation}

Subsequently, a pretrained text decoder $\mathbf{D}$ transforms the poisoned embedding $x^p_i$ into natural-language text:
\begin{equation}
    \tilde{t}_i = \mathbf{D}(x^p_i).
\end{equation}

To further improve semantic coherence, TextTrojan is jointly optimized with a shadow GNN $f_{\psi}$ so that the generated poison text moves poisoned nodes toward the target class while preserving clean-node behavior. We study two ways of injecting the generated trigger into the original node text: \textit{Overwriting} and \textit{Appending}.

The \textbf{Overwriting} strategy replaces the original node text entirely with the generated poisoned text. This gives the attacker stronger control over the final semantics and usually leads to higher ASR, but it may also increase the risk of detectable linguistic anomalies:
\begin{equation}
    t^p_i = \tilde{t}_i.
\end{equation}

By contrast, the \textbf{Appending} strategy retains the original text and appends the generated trigger text at the end, thereby preserving more of the original semantics and linguistic coherence. This makes the poison text harder to notice, although it usually requires stronger optimization to ensure that the trigger still dominates the prediction:
\begin{equation}
    t^p_i = t_i \oplus \tilde{t}_i,
\end{equation}
where $\oplus$ denotes textual concatenation.

After text perturbation, all final poisoned texts are re-encoded into vector form $\tilde{X}_p$ by a pretrained text encoder $\mathbf{E}$ for downstream GNN training. The overall optimization objective balances attack effectiveness, clean classification accuracy, and semantic similarity:
\begin{equation}
\begin{aligned}
    \min_{\phi} \Big(
    & \mathcal{L}_{\text{atk}}(\phi; f_{\psi}, X_p, y_t) \\
    & + \mathcal{L}_{\text{clean}}(\phi; f_{\psi}, X, y) \\
    & + \lambda\mathcal{L}_{\text{sim}}(X_p,\tilde{X}_p)
    \Big),
\end{aligned}
\end{equation}
where $\phi$ denotes parameters of the poisoned text generator, $X$ represents the original text embeddings of labeled nodes, $X_p$ and $\tilde{X}_p$ represent embeddings of original and poisoned texts of poisoned nodes respectively, $y_t$ denotes the targeted poisoned class, and $\lambda$ balances semantic similarity with attack effectiveness.

The attack effectiveness loss $\mathcal{L}_{\text{atk}}$ ensures poisoned nodes are classified into the target label $y_t$:
\begin{equation}
    \mathcal{L}_{\text{atk}}(\phi; f_{\psi}, X_p, y_t) = -\sum_{v_i \in \mathcal{V}_p}\log f_{\psi}(y_t|x^p_i;\phi).
\end{equation}

The clean classification loss $\mathcal{L}_{\text{clean}}$ ensures minimal interference with clean node classification:
\begin{equation}
    \mathcal{L}_{\text{clean}}(\phi; f_{\psi}, X, y) = -\sum_{v_i \in \mathcal{V}_L}\log f_{\psi}(y|x_i;\phi),
\end{equation}
where $\mathcal{V}_L$ denotes labeled nodes.

The semantic similarity loss $\mathcal{L}_{\text{sim}}$ employs cosine embedding similarity to maintain textual naturalness:
\begin{equation}
    \mathcal{L}_{\text{sim}}(X_p,\tilde{X}_p)=\sum_{v_i \in \mathcal{V}_p}\left(1-\cos(\mathbf{x}(t_i),\mathbf{x}(\tilde{t}_i))\right).
\end{equation}

This optimization matches the design goal introduced above. The attack loss makes the trigger effective, the clean loss keeps the poisoned model useful on benign data, and the similarity loss discourages obviously abnormal text. Together, these terms allow TAGBD to generate poison text that adapts to graph context while remaining relatively inconspicuous.

From a practical perspective, the two injection modes correspond to different attacker priorities. Overwriting is suitable when the attacker has strong control over the visible text of a poisoned node and cares primarily about maximizing the learned trigger-target association. Appending is suitable when the attacker wants the final text to retain most of its original meaning and surface form, for example when profile descriptions, abstracts, or short posts may later be inspected by moderators. Evaluating both modes therefore helps us understand not only the strongest threat, but also the more stealth-oriented threat that is likely to matter in practice.

It is also worth emphasizing that TAGBD does not depend on a single fixed trigger phrase shared across all nodes. Instead, the generated poison text is instance-dependent because it is derived from graph-aware node representations. This design makes the attack more realistic and harder to detect. A static trigger can often be exposed by simple frequency analysis or manual inspection, especially when the poisoned budget is small. In contrast, an instance-aware trigger family can preserve local semantic variation while still encoding a consistent signal toward the target label. This is one of the main reasons why graph-aware text generation is a more powerful threat model than naive textual insertion.

\subsection{Overall Attack Procedure}
To summarize how the above components interact, Algorithm~\ref{alg:tagbd} presents the full poisoning procedure. At a high level, TAGBD first identifies the nodes that are most amenable to poisoning, then learns stealthy trigger text for those nodes, and finally uses the poisoned graph to train the target model. This ordering follows the logic of the running example and connects node selection with trigger generation in one end-to-end workflow.

\begin{algorithm}[ht]
\caption{Joint Training-Based Textual Backdoor Injection}
\label{alg:tagbd}
\begin{algorithmic}[1]
\Require Graph $\mathcal{G} = (\mathcal{V}, \mathcal{E}, \mathbf{X})$, labels $Y$, poisoning ratio $\gamma$, target label $y_{\text{target}}$, trigger strategy \texttt{mode} $\in \{$overwrite, append$\}$
\Ensure Poisoned training set and final GNN model $\hat{f}_\theta$

\State \textbf{Initialize:} surrogate GNN $\hat{f}_p$, TextTrojan generator $\mathcal{T}_t$, text encoder $\mathbf{E}$, text decoder $\mathbf{D}$
\State Train $\hat{f}_p$ and $\mathcal{T}_t$ jointly on labeled nodes $\mathcal{V}_L$ and poisoned nodes $\mathcal{V}_p$
\State Compute uncertainty $H(v)$ for all unlabeled nodes $v \in \mathcal{V}_U$
\State Select top-$\gamma \cdot |\mathcal{V}|$ uncertain nodes as $\mathcal{V}_p$

\While{target GNN model $\hat{f}_\theta$ not converged}
\ForAll{$v \in \mathcal{V}_p$}
    \State Obtain node embedding $\mathbf{z}_v = \hat{f}_p(v)$
    \State Obtain neighbor embeddings $\{\mathbf{z}_u\ |\ u \in \mathcal{N}(v)\}$
    \State $\hat{\mathbf{x}}_v \gets \mathcal{T}_t(\mathbf{z}_v, \{\mathbf{z}_u\})$
    \State $t^{\text{trig}}_v \gets \mathbf{D}(\hat{\mathbf{x}}_v)$
    \If{\texttt{mode == overwrite}}
        \State $t^{\text{p}}_v \gets t^{\text{trig}}_v$
    \Else
        \State $t^{\text{p}}_v \gets t^{\text{orig}}_v \oplus t^{\text{trig}}_v$
    \EndIf
    \State $x^{\text{poison}}_v \gets \mathbf{E}(t^{\text{p}}_v)$
    \State Replace $x_v \in \mathbf{X}$ with $x^{\text{poison}}_v$, and assign label $y_{\text{target}}$
    \State Train target GNN model $\hat{f}_\theta$
\EndFor
\EndWhile

\State \Return $\hat{f}_\theta$, $\mathcal{T}_t$
\end{algorithmic}
\end{algorithm}

Algorithm~\ref{alg:tagbd} can be read as a three-stage attack pipeline. In the first stage, TAGBD trains a surrogate model and uses it to identify a small set of vulnerable nodes whose labels are easier to redirect. In the second stage, the TextTrojan generator converts graph-aware node embeddings into poisoned text embeddings and decodes them into natural-language triggers. In the third stage, these poisoned texts are injected back into the training graph and used to train the final target model. The final backdoor therefore arises from poisoned supervision rather than from any architectural change to the target GNN.

This staged design has two practical advantages. First, it cleanly separates \emph{selection}, \emph{generation}, and \emph{implantation}, which makes it easier to understand where the attack obtains its effectiveness. Selection improves budget efficiency, generation improves the quality and contextual fit of the poison text, and implantation writes the trigger-target association into the final training distribution. Second, the attack remains modular with respect to the deployed model. The surrogate and shadow modules are tools for constructing the poisoned data; they do not need to match the final GNN exactly for the attack to transfer.

From the defender's perspective, this workflow also explains why the attack is difficult to remove after training. Once the poisoned graph has been used to fit the target model, the malicious behavior is embedded in the learned association between text patterns, graph context, and the target label. At that point, removing only a few suspicious nodes may no longer be sufficient, especially if the poison text was designed to remain semantically plausible and distributed across different classes. This persistence is one of the reasons why training-time backdoors remain especially concerning in practice.

\section{Experiments}\label{se:experiments}
This section evaluates TAGBD from four complementary perspectives: auxiliary analyses that explain how the attack behaves, main comparisons against strong baselines, transferability across target models and defenses, and component-level ablation. Following the principle of ``first support, then main results,'' we begin with experiments that clarify poisoning budgets, text generators, and trigger lengths, and then move to the primary performance comparisons.

\subsection{Experimental Settings}

\paragraph{Datasets.} Experiments are conducted on three widely used TAG benchmarks from TAGLAS \cite{taglas}: Cora, Pubmed, and Arxiv. Table~\ref{tab:datasets} summarizes their basic statistics.

\begin{table}[h]
\centering
\caption{Statistics of the three evaluation datasets. Words$_{avg}$ and Words$_{max}$ denote the average and maximum number of words per node text.}
\label{tab:datasets}
\resizebox{\linewidth}{!}{
\begin{tabular}{lccccc}
    \toprule
    \textbf{Datasets} & \textbf{Nodes} & \textbf{Edges} & \textbf{Classes} & \textbf{Words$_{avg}$} & \textbf{Words$_{max}$} \\
    \midrule
    Cora & 2,708 & 10,556 & 7  & 138 &763\\ 
    Pubmed & 19,717 & 88,648 &  3  & 247&895\\
    ArXiv & 169,343 & 1,166,243 &  40  & 421 &1414\\ 
    \bottomrule
\end{tabular}
}
\end{table}

Table~\ref{tab:datasets} shows that the evaluation spans a broad difficulty range rather than a single narrow setting. Cora is relatively small and suitable for controlled comparison, Pubmed contains longer texts and a medium-scale graph, and ArXiv introduces a much larger graph with richer class diversity. This progression matters because a convincing TAG backdoor should remain effective not only on small citation benchmarks, but also when graph scale, text length, and label complexity increase simultaneously.

\paragraph{Baselines, Defenses, and Target Models.}
TAGBD is compared with five representative graph backdoor attacks. SBA-Samp and SBA-Gen~\cite{sba2021} poison the graph through static or generated subgraphs. GTA~\cite{gta2021} learns sample-specific structural triggers. UGBA~\cite{ugba2023} emphasizes representative poisoned nodes and homophily-preserving triggers. DPGBA~\cite{dpgba2024} further improves stealth through in-distribution adversarial training. Together, these baselines cover the most influential structural backdoor paradigms and therefore provide a strong reference point for evaluating a text-only attack.
To assess robustness under defenses, we adopt Prune~\cite{ugba2023}, which removes suspicious low-similarity edges, and OD~\cite{dpgba2024}, which detects and filters structural outliers. These defenses are especially informative here because they target topology-oriented triggers. If TAGBD remains effective after they are applied, then text-only poisoning is exploiting a vulnerability that current structural defenses do not fully cover.
To assess transferability, poisoned graphs are trained and tested on GCN~\cite{gcn2017}, GraphSAGE~\cite{graphsage2017}, GAT~\cite{gat2017}, GraphTransformer, and RobustGCN~\cite{robustgcn2019}.

\paragraph{Evaluation Metrics.}
TAGBD is evaluated on inductive node classification, where the attacker has no access to test nodes during poisoning. To measure both effectiveness and stealth, we use the following three metrics:

\begin{itemize}[leftmargin=10pt,topsep=0pt,partopsep=0pt]
    \item \textit{Attack Success Rate (ASR):} The proportion of test nodes embedded with the trigger that are misclassified into the attacker-specified target class. ASR reflects the effectiveness of the backdoor.
    
    \item \textit{Clean Accuracy (CA).} The classification accuracy of the backdoored GNN on clean (unmodified) test nodes. High CA indicates that the backdoor injection does not impair the model’s performance on benign data, preserving its utility.
    
    \item \textit{Perplexity (PPL).} To quantify the linguistic naturalness of the poisoned texts, we use the standard \textit{perplexity} metric. PPL measures how well a language model predicts the next token in a sequence and serves as a proxy for human-perceived fluency. Given a tokenized text sequence \( t = t_0, t_1, \dots, t_{n-1} \), perplexity is defined as:
    \[
    \text{PPL}(t) = \exp\left(-\frac{1}{n-1} \sum_{i=1}^{n-1} \log p(t_i \mid t_0, \dots, t_{i-1})\right)
    \]
\end{itemize}

A strong attack should achieve high ASR while maintaining high CA and low PPL. These three metrics jointly capture targeted control, preserved model utility, and the stealth of the injected poison text.

\paragraph{Implementation Details.}
For each dataset, 20\% of nodes are randomly split into the test set, and that test set is further divided evenly into targeted and clean subsets. The remaining 80\% of nodes are used for training, with 10\% labeled nodes and 10\% validation nodes. The poisoning budget is fixed at 1\% of unlabeled training nodes unless otherwise stated.

For TAGBD, a 2-layer GCN is used as the surrogate/shadow model with hidden dimension 512, and a 2-layer MLP is used as the trigger generator with hidden dimension 1024. The learning rate is 0.01, weight decay is 0.0005, dropout is 0.5, the maximum epoch number is 300, and the early-stopping patience is 50. The semantic similarity weight is set to $\lambda = 0.5$ and tuned from \(\{0.1, 0.5, 0.9\}\). The maximum poisoned text length is 1024 tokens for overwriting and 512 tokens for appending. Unless otherwise stated, all methods use Sonar text embeddings~\cite{Duquenne:2023:sonar_arxiv}.

All experiments are repeated five times with different random seeds, and average results are reported. Experiments are run on a Linux server with an Intel(R) Xeon(R) Gold 6348 CPU at 2.60GHz, 1.0~TiB RAM, and an NVIDIA A800-SXM4 GPU with 80~GB memory.
For reproducibility, the source code is publicly available at \url{https://github.com/iicalgorithms/TAGBackdoor.git}.

This setup is intentionally conservative in one important sense: the attacker does not poison a large fraction of the graph and does not rely on access to the test set. The evaluation therefore focuses on whether a small amount of training-time poison text can implant a persistent and reusable backdoor. This setting is appropriate for TAG applications in which content can be manipulated incrementally over time, but full control over the data pipeline is unrealistic.

\begin{table*}[ht]
\caption{Main comparison with graph backdoor baselines under different defense settings. `Clean' reports the accuracy of the benign model. For each attack method, the left entry is ASR and the right entry is CA. The last two groups correspond to TAGBD with overwriting and appending triggers; boldface marks the best ASR in each row.}
\label{tab:compare_baselines}
\resizebox{\textwidth}{!}{
\begin{tabular}{c|c|c|rl|rl|rl|rl|rl|rl|rl}
\toprule
\textbf{Datasets}       & \textbf{Defense} & \textbf{Clean} & \multicolumn{2}{c}{\textbf{SBA-Samp}} & \multicolumn{2}{c}{\textbf{SBA-Gen}} & \multicolumn{2}{c}{\textbf{GTA}} & \multicolumn{2}{c}{\textbf{UGBA}} & \multicolumn{2}{c}{\textbf{DPGBA}} & \multicolumn{2}{c}{\textbf{Overwriting}} & \multicolumn{2}{c}{\textbf{Appending}} \\
\midrule
\multirow{3}{*}{Cora}   & None             & 77.82          & 27.23             & 76.52             & 42.23             & 77.12            & 92.55           & 79.93          & 95.25           & 79.22           & 93.25            & 78.22           & \textbf{100.00}       & 86.72      & \textbf{99.63}        & 86.72      \\
                        & Prune            & 77.23          & 17.25             & 75.93             & 21.35             & 78.62            & 0.25            & 78.61          & 93.33           & 78.38           & 19.47            & 77.98           & \textbf{100.00}       & 87.82      & \textbf{99.37}       & 85.61      \\
                        & OD               & 76.92          & 30.50             & 75.62             & 43.37             & 77.94            & 46.12           & 78.24          & 0.00            & 78.32           & 93.94            & 77.92           & \textbf{100.00}       & 87.45      & \textbf{95.20}        & 91.14      \\
\midrule
\multirow{3}{*}{Pubmed} & None             & 81.56          & 28.83             & 83.41             & 30.51             & 85.12            & 95.76           & 82.22          & 92.15           & 83.96           & 92.55            & 80.26           & \textbf{99.59}        & 93.05      & \textbf{98.59}        & 93.05      \\
                        & Prune            & 81.61          & 25.17             & 82.56             & 22.73             & 84.62            & 26.31           & 82.73          & 91.84           & 84.71           & 89.59            & 80.61           & \textbf{99.85}        & 93.10      & \textbf{98.90}        & 91.18      \\
                        & OD               & 83.69          & 23.57             & 82.74             & 29.63             & 83.94            & 87.34           & 82.49          & 30.25           & 83.99           & 18.84            & 82.83           & \textbf{99.85}        & 92.75      & \textbf{97.92}        & 94.17      \\
\midrule
\multirow{3}{*}{arxiv}  & None             & 57.27          & 19.62             & 58.24             & 50.67             & 57.12            & 74.16           & 58.77          & 97.92           & 57.53           & 96.64            & 56.25           & \textbf{99.52}        & 59.74      & 96.47                 & 64.71      \\
                        & Prune            & 57.00          & 0.25              & 53.51             & 0.26              & 56.63            & 0.00            & 55.00          & 96.62           & 58.49           & 0.00             & 58.30           & \textbf{99.96}        & 73.00      & \textbf{97.70}        & 65.13      \\
                        & OD               & 56.96          & 15.26             & 55.84             & 41.56             & 56.52            & 0.00            & 56.96          & 15.22           & 58.96           & 94.17            & 57.46           & \textbf{99.73}        & 59.80      & 81.61                 & 70.46     
\\
\bottomrule
\end{tabular}
}
\end{table*}

\begin{table}[ht]
\caption{Transferability across target GNNs on the ArXiv dataset. For each strategy, the left entry is ASR and the right entry is CA.}
\label{tab:compare_gnns}
\centering
\begin{tabular}{c|cc|cc}
\toprule
\textbf{Target Model}     &  \multicolumn{2}{c}{\textbf{Overwriting}} & \multicolumn{2}{c}{\textbf{Appending}} \\
\midrule
GCN                  & 99.25          & 86.35         & 97.86         & 63.86         \\
GraphSAGE             & 99.62          & 83.03         & 98.99         & 69.83         \\
GAT                   & 99.26           & 83.76         & 98.89         & 61.55         \\
GraphTransformer     & 99.17          & 85.24         & 99.15         & 68.19         \\
RobustGCN            & 98.89           & 87.45         & 91.94         & 66.88         \\
\bottomrule
\end{tabular}
\end{table}

\subsection{Auxiliary Analyses}

\begin{figure}[t]
    \centering
    \includegraphics[width=0.45\textwidth]{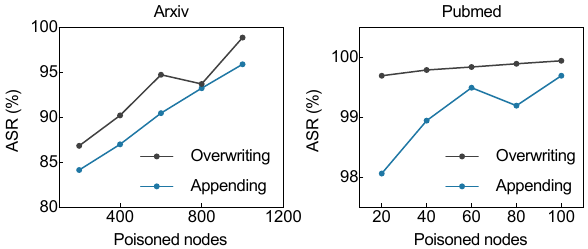}
    \caption{Attack success rate under different poisoning budgets. Both variants improve as more training nodes are poisoned, and overwriting reaches strong performance with fewer poisoned nodes.}
    \label{fig:poisoned_nodes}
\end{figure}

\paragraph{Effect of Number of Poisoned Nodes.}
To investigate sensitivity to poisoning budget, we evaluate the impact of varying numbers of poisoned nodes on the attack success rate. Since earlier experiments show that using only 1\% poisoned nodes (approximately 27 nodes) already achieves strong attack performance on Cora, this analysis focuses on the larger-scale datasets ArXiv and Pubmed.
Specifically, we evaluate poisoned-node counts of \{20, 40, 60, 80, 100\} for Pubmed, and \{200, 400, 600, 800, 1000\} for ArXiv.
Clean accuracy remains largely stable under different poisoning ratios, so we primarily report ASR in Fig.~\ref{fig:poisoned_nodes}.
For both ArXiv and Pubmed, ASR increases consistently as the number of poisoned nodes grows, reflecting greater attack leverage under larger poisoning budgets. The overwriting strategy maintains superior performance across all tested poisoning levels and reaches nearly perfect ASR even with relatively few poisoned nodes, around 20 nodes for Pubmed and 200 nodes for ArXiv. By comparison, the appending strategy exhibits slightly lower ASR at smaller poisoning scales but gradually approaches the performance of overwriting as the number of poisoned nodes grows.
These observations indicate that TAGBD remains highly effective even when the attacker can poison only a very small portion of the training set. This result strengthens the realism of the threat model, because the attack does not depend on an unrealistically large poisoning budget.
An equally important implication is that the learned trigger-target association is data-efficient. The attack does not require thousands of poisoned examples to dominate training. Instead, a carefully chosen and graph-aware set of poisoned nodes is enough to bias the learned decision rule. For security evaluation, this is a more concerning scenario than large-budget poisoning because it is easier to carry out gradually and less likely to be noticed during data curation.

\paragraph{Effect of Text Encoder-Decoder Models.}
To examine how different text encoder-decoder architectures affect attack effectiveness and stealth, we evaluate the poisoning strategies using three widely adopted text representation models: Bag-of-Words (BOW)~\cite{bow}, GTR-T5~\cite{vec2text}, and Sonar~\cite{Duquenne:2023:sonar_arxiv}. The dimensionality of text embeddings is set to the default configuration (BOW=1024, GTR-T5=768, Sonar=1024).
Results on the Cora dataset are reported in Table~\ref{tab:encoder_decoder_comparison}.
%



\begin{table}[t]
\centering
\caption{Effect of encoder-decoder choice on Cora. Each entry reports ASR, CA, and PPL. Stronger text models generally improve both attack effectiveness and text quality.}
\label{tab:encoder_decoder_comparison}
\begin{tabular}{l|ccc|ccc}
\toprule
\multirow{2}{*}{\textbf{Model}} 
& \multicolumn{3}{c|}{\textbf{Overwriting}} 
& \multicolumn{3}{c}{\textbf{Appending}} \\
& ASR & CA & PPL & ASR & CA & PPL \\
\midrule
BOW    & 50.92 & 88.39 & 211.73 & 18.82  & 86.35 & 220.33 \\
GTR-T5  & 98.52 & 87.08 & 341.62 & 95.57 & 86.72 & 62.18 \\
Sonar  & 100.00 & 87.22 & 175.23 & 99.96 & 76.35 & 95.45 \\
\bottomrule
\end{tabular}
\end{table}

Table~\ref{tab:encoder_decoder_comparison} shows that the choice of text encoder-decoder is not a minor implementation detail, but a central factor in the success of the attack. BOW performs poorly in both effectiveness and fluency, which suggests that weak text representations cannot encode a strong trigger while preserving natural language. GTR-T5 improves this balance substantially, especially for appending. Sonar performs best overall: it reaches the highest ASR under both strategies and maintains a more favorable perplexity profile than BOW. This result supports the main intuition behind TAGBD, namely that graph-aware trigger optimization is most effective when the text backbone is strong enough to map optimized embeddings back into plausible text.

This comparison is useful beyond implementation selection. It also clarifies what makes stealthy poison text dangerous. The risk is not merely that text can be modified, but that modern text models can encode a malicious signal without obviously sacrificing fluency. As text encoders and decoders improve, the practical barrier to generating believable poisoned content decreases. In that sense, stronger language models can amplify the backdoor risk of TAG systems unless defenses also become content-aware.

\subsection{Impact of Poisoned Text Length}

\begin{figure}
    \centering
    \includegraphics[width=1\linewidth]{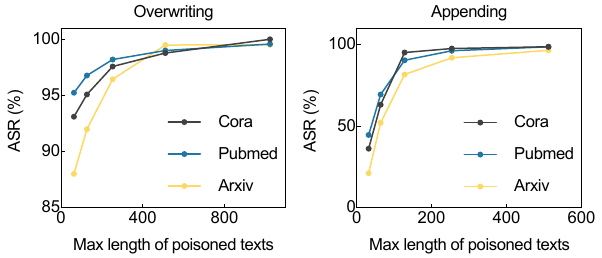}
    \caption{Effect of trigger length on attack success rate. Longer triggers improve ASR for both variants, while overwriting reaches saturation earlier than appending.}
    \label{fig:text_length_effect}
\end{figure}

To investigate how poisoned text length affects attack performance, we systematically vary the maximum allowed trigger length for both injection strategies. For overwriting, we directly adjust the maximum length of poisoned texts and set the length to \{64, 128, 254, 512, 1024\}. For appending, we vary only the length of the appended trigger phrase and select lengths from \{32, 64, 128, 254, 512\}. All other experimental parameters, including the number of poisoned nodes, encoder-decoder models, and training settings, remain fixed.
Experimental results are summarized in Fig.~\ref{fig:text_length_effect}. The trend is consistent across datasets: ASR increases with trigger length for both strategies. In particular, the overwriting strategy approaches saturation at around 512 tokens, indicating that strong attacks can already be achieved with moderately sized poisoned texts.
By contrast, the appending strategy initially exhibits lower ASR at very short trigger lengths (32--64 tokens). However, ASR increases sharply and approaches saturation once the trigger length reaches 128 tokens and beyond, eventually achieving performance comparable to overwriting.
Overall, these findings confirm that longer poisoned texts generally facilitate stronger attacks, but they also show that TAGBD does not require unrealistically long triggers to perform well in practice. This observation further supports the plausibility of the attack in realistic settings where very long or obviously repetitive text would be easy to flag.

The length study also helps interpret the difference between overwriting and appending. Overwriting reaches saturation earlier because the entire node text is under the attacker's control, so each additional token directly contributes to the poisoned semantics. Appending behaves differently: it must compete with the original text that remains present in the node, which means the generated trigger has to be long enough to become influential without becoming visually suspicious. This tension is exactly the kind of effectiveness-stealth trade-off that practical defenders should care about.



\begin{table*}[ht]
\centering
\small
\caption{Case study of poisoned texts generated by overwriting and appending when Sonar is used as the text encoder-decoder.}
\label{tab:poisoned_texts_sonar}
\begin{tabular}{>{\raggedright\arraybackslash}p{0.58\linewidth}|>{\raggedright\arraybackslash}p{0.17\linewidth}|>{\raggedright\arraybackslash}p{0.17\linewidth}}
\toprule
Original label: \textbf{Rule Learning}  & \multicolumn{2}{|l}{Target label: \textbf{Case-Based}} \\
\toprule
Original text  & Overwriting      & Appending     \\
\midrule
 Extraction of meta-knowledge to restrict the hypothesis space for ILP systems. Many ILP systems, such as GOLEM, FOIL, and MIS, take advantage of user supplied meta-knowledge to restrict the hypothesis space. This meta-knowledge can be in the form of type information about arguments in the predicate being learned, or it can be information about whether a certain argument in the predicate is functionally dependent on the other arguments (supplied as mode information). This meta knowledge is explicitly supplied to an ILP system in addition to the data. The present paper argues that in many cases the meta knowledge can be extracted directly from the raw data. Three algorithms are presented that learn type, mode, and symmetric meta-knowledge from data. These algorithms can be incorporated in existing ILP systems in the form of a preprocessor that obviates the need for a user to explicitly provide this information. In many cases, the algorithms can extract meta- knowledge that the user is either unaware of, but which information can be used by the ILP system to restrict the hypothesis space.              
 &  In a nutshell, there are dozens of arguments for the elimination of algorithms, dozens of arguments for the elimination of algorithms, dozens of arguments for the elimination of algorithms, dozens of arguments for the elimination of algorithms, dozens of arguments for the elimination of algorithms.               
 &  In the academic literature, a number of metaphors, titles, and an introductory introduction, knowledge of theoretical knowledge in order to eliminate the limitations of the algorithm in order to allow for the possibility of ignoring the algorithm.            
 \\
\bottomrule
\end{tabular}
\end{table*}

\begin{table*}[ht]
\centering
\small
\caption{Case study of poisoned texts generated by overwriting and appending when BOW is used as the text encoder-decoder.}
\label{tab:poisoned_texts_bow}
\begin{tabular}{>{\raggedright\arraybackslash}p{0.38\linewidth}|>{\raggedright\arraybackslash}p{0.37\linewidth}|>{\raggedright\arraybackslash}p{0.17\linewidth}}
\toprule
Original label: \textbf{Neural Networks}  & \multicolumn{2}{|l}{Target label: \textbf{Case-Based}} \\
\toprule
Original text  & Overwriting      & Appending     \\
\midrule
 Adaptive parameter pruning in neural networks. Neural network pruning methods on the level of individual network parameters (e.g. connection weights) can improve generalization. An open problem in the pruning methods known today (OBD, OBS, autoprune, epsiprune) is the selection of the number of parameters to be removed in each pruning step (pruning strength). This paper presents a pruning method lprune that automatically adapts the pruning strength to the evolution of weights and loss of generalization during training. The method requires no algorithm parameter adjustment by the user. The results of extensive experimentation indicate that lprune is often superior to autoprune (which is superior to OBD) on diagnosis tasks unless severe pruning early in the training process is required. Results of statistical significance tests comparing autoprune to the new method lprune as well as to backpropagation with early stopping are given for 14 different problems.              
 & user neural of e results pruning to to results individual improve the that the by the early evolution backpropagation as early parameter the to often is is parameters and for to open tests of training the pruning in is of with in method in be number paper no on method of the neural pruning is with as pruning each in requires diagnosis well pruning pruning are of pruning comparing method selection of required can tasks the in early step and parameter problem the during pruning level the the adaptive parameters pruning to indicate to methods a g the parameters in method is as automatically of algorithm generalization that network to statistical to on methods network academic different of known loss of problems this title the networks the paper the results weights weights which abstract of presents process with pruning training new an the generalization the given pruning.
 &  to pruning to the the in the method is to pruning of weights of with and pruning of results the, ilp user as in the ilp systems algorithms the the from information the the space that the the that knowledge of many in to knowledge and this can can be knowledge hypothesis is be data a system in to of can information, <UNK>, to of based to of a we approach in and.    
 \\
\bottomrule
\end{tabular}
\end{table*}

\subsection{Main Attack Performance}
We next report the primary evaluation results. Building on the auxiliary analyses above, this subsection compares TAGBD with strong baselines, evaluates transferability across target models, and examines the linguistic naturalness of the generated texts.

\paragraph{Comparison with Baseline Attacks.}
Table~\ref{tab:compare_baselines} reports the main comparison under different defense settings.

\paragraph{Transferability Across Target Models.}
Table~\ref{tab:compare_gnns} reports how the learned backdoor transfers across different target architectures on ArXiv.

\begin{figure}
    \centering
    \includegraphics[width=1\linewidth]{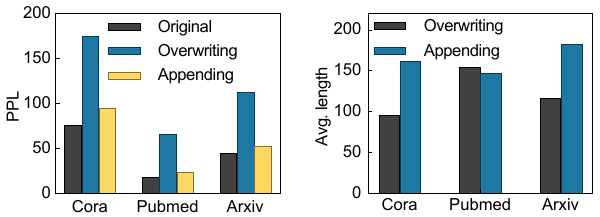}
\caption{Perplexity and average text length under different poisoning strategies. Overwriting yields stronger but less natural text, while appending better preserves the linguistic properties of clean samples. Lower perplexity indicates more natural text.}
    \label{fig:ppl}
\end{figure}

\paragraph{Stealthiness of Poisoned Texts.}
Figure~\ref{fig:ppl} evaluates the linguistic side of the attack through perplexity and average text length. Overwriting produces stronger but less natural triggers because it replaces the original text more aggressively, which raises perplexity and shortens the resulting content. Appending preserves more of the original text, so its perplexity and length distribution remain closer to those of clean texts. This result matches the intended trade-off between the two variants: overwriting prioritizes attack strength, whereas appending provides a more balanced effectiveness-stealth profile.

Table~\ref{tab:compare_baselines} provides the central empirical result of the paper. Across all three datasets and nearly all defense settings, TAGBD achieves the highest ASR, often by a large margin. The trend is particularly clear on Cora and Pubmed, where the overwriting variant remains at or near 100\% ASR even after Prune or OD is applied, while the appending variant also remains highly effective. On ArXiv, which is much larger and more challenging, TAGBD still outperforms prior methods and maintains very high ASR. The clean accuracy values are equally important: TAGBD does not succeed by collapsing the classifier. Instead, it preserves competitive CA while sharply increasing targeted misclassification, indicating that the learned model remains useful on benign inputs but becomes selectively controllable on triggered inputs.

The results are especially revealing when read row by row against the defenses. Structural attacks often lose much of their ASR after Prune or OD because their triggers are encoded in graph patterns that those defenses are explicitly designed to suppress. TAGBD behaves differently. Since the attack leaves graph topology untouched, the defense is forced to act on a signal that is largely orthogonal to its design assumptions. This mismatch explains why text-only poisoning remains strong even when structural defenses are active. From a security perspective, the lesson is clear: a defense that focuses only on the graph channel may leave the text channel almost entirely unmonitored.

There is also an important dataset-level pattern. On Cora and Pubmed, TAGBD reaches near-saturated ASR under both injection modes, indicating that the backdoor is easy to implant even when the graph is relatively modest in scale. On ArXiv, although the task is harder due to graph size and label diversity, TAGBD still preserves a very strong trigger-target effect. This suggests that the attack is not an artifact of small citation benchmarks. Instead, the vulnerability persists as the TAG becomes larger and semantically richer, which strengthens the practical relevance of the threat model.

Table~\ref{tab:compare_gnns} further shows that the learned backdoor is not tied to a single deployment architecture. Under overwriting, ASR stays around 99\% for all tested models, including RobustGCN, which is designed to be more robust. Appending is slightly weaker, but still transfers strongly across GCN, GraphSAGE, GAT, GraphTransformer, and RobustGCN. This pattern suggests that the attack is not simply overfitting to one surrogate-target pairing. Rather, the poisoned training data induces a stable trigger-target association that survives architectural variation.

This cross-model transfer is important because real attackers rarely know the exact model that will be deployed. A text-only backdoor would be far less concerning if it required tight white-box alignment between the poisoning pipeline and the final classifier. Our results suggest the opposite. Once the trigger-target association is successfully written into the poisoned training distribution, different downstream GNNs can absorb that association even when they vary in message-passing mechanism or robustness design. This observation makes the threat model meaningfully stronger.

Taken together with Fig.~\ref{fig:ppl}, the main experiments support three conclusions. TAGBD is effective, it transfers across models, and it exposes a vulnerability that existing topology-oriented defenses do not adequately address. The comparison between overwriting and appending is also consistent throughout the section: overwriting offers maximum attack strength, whereas appending yields a more favorable effectiveness-stealth balance.

\subsection{Discussion and Security Implications}
The experimental results support a broader interpretation of TAG security. In many deployed systems, defenses and monitoring tools are still organized around graph artifacts, such as suspicious edges, dense inserted communities, or anomalous subgraph motifs. TAGBD shows that this is only part of the picture. A determined attacker may leave the graph untouched and instead use poison text as the carrier of malicious behavior. In such a scenario, topology-only defenses can report a graph that appears normal even though the learned model has already been backdoored.

The overwriting and appending variants also help clarify attacker capabilities. Overwriting represents the stronger but more intrusive adversary, for example one who can fully control the text attached to selected training nodes. Appending represents a weaker but stealthier adversary who must preserve the original content while still inserting a hidden cue. The fact that both variants remain highly effective across datasets suggests that the practical attack surface is broad: defenders cannot assume safety merely because full text replacement is implausible in their application.

Finally, the experiments point toward a new direction for defense design. If the core vulnerability arises from the interaction between graph propagation and node content, then defenses should evaluate those two channels jointly. Promising directions may include content-aware outlier detection, consistency checks between node text and neighborhood semantics, and training-time auditing that measures whether small textual cues disproportionately influence the final prediction. We do not claim that the current paper solves the defense problem, but the results make clear that such defenses are urgently needed.

\subsection{Qualitative Case Study}
The perplexity results above show that the two trigger injection strategies differ in linguistic naturalness, but they do not reveal how the poisoned texts appear in practice. To make this distinction concrete, we provide qualitative examples in Tables~\ref{tab:poisoned_texts_sonar} and~\ref{tab:poisoned_texts_bow}. These examples illustrate the kinds of poisoned text that a defender or reviewer would observe.

Taken together, Tables~\ref{tab:poisoned_texts_sonar} and~\ref{tab:poisoned_texts_bow} make the difference in stealth visually apparent. Sonar-generated texts remain more fluent and contextually grounded after poisoning, whereas BOW outputs often degenerate into disordered keyword sequences. The appending strategy is also visibly more natural because it preserves substantially more of the original content. This qualitative evidence is consistent with the quantitative conclusions from Table~\ref{tab:encoder_decoder_comparison} and Fig.~\ref{fig:ppl}: stronger text generation backbones and less destructive injection strategies lead to more realistic poison text.

The qualitative examples also suggest why linguistic evaluation should not be treated as secondary in TAG backdoor research. A trigger that achieves high ASR but produces visibly broken text may be easy to remove during content moderation or data cleaning. By contrast, a trigger that blends into the surrounding semantics can survive much longer in real-world pipelines. In this sense, stealth is not merely an aesthetic property of the poisoned text; it is a core factor that determines whether the backdoor can persist until deployment.

\subsection{Ablation Study}
To assess the contributions of the key components in TAGBD, we conduct an ablation study focusing on three modules: (1) the poisoned node selection module (NS), (2) the GNN-based node embedding module (NE), and (3) the feature similarity constraint (FS) between poisoned and original text embeddings.
We evaluate three variants of TAGBD. For NS, uncertainty-guided selection is replaced with random sampling of poisoned nodes from the training graph. For NE, the GNN-derived node embeddings are replaced with direct text embeddings when training the TextTrojan trigger generator. For FS, the semantic similarity loss is removed so that the model optimizes triggers without preserving the embedding similarity between poisoned and original texts.


\begin{table}[t]
\centering
\caption{Ablation study on ArXiv under both overwriting and appending strategies. Each entry reports ASR, CA, and PPL.}
\label{tab:ablation_study}
\begin{tabular}{l|ccc|ccc}
\toprule
\multirow{2}{*}{\textbf{Variant}} 
& \multicolumn{3}{c|}{\textbf{Overwriting}} 
& \multicolumn{3}{c}{\textbf{Appending}} \\
& ASR & CA & PPL & ASR & CA & PPL \\
\midrule
Full TAGBD    & 99.7 & 60.52 & 124.2 & 93.2 & 64.6 & 101.8 \\
w/o NS    & 89.3 & 64.4 & 120.5 & 79.4 & 64.7 & 99.7 \\
w/o NE   & 85.5 & 64.9 & 118.2 & 76.3 & 64.8 & 97.1 \\
w/o FS   & 98.2 & 63.1 & 228.6 & 91.6 & 63.7 & 153.5 \\
\bottomrule
\end{tabular}
\end{table}

Table~\ref{tab:ablation_study} shows that the gains of TAGBD do not come from a single isolated component. Removing NS causes a clear drop in ASR under both strategies, which confirms that uncertainty-guided node selection is important when the poisoning budget is limited. Removing NE further reduces ASR, indicating that graph-aware node embeddings provide useful structural information that plain text embeddings miss. By contrast, removing FS leaves ASR relatively high but sharply increases perplexity, especially under overwriting. This separation is informative: NS and NE mainly support attack effectiveness, whereas FS mainly supports linguistic naturalness. The full model performs best because these components solve different parts of the problem rather than duplicating one another.

The ablation results also reinforce the main conceptual claim of the paper. TAGBD is not effective simply because it adds extra optimization capacity to the poisoning pipeline. If that were the case, removing any single component would have only a minor or inconsistent effect. Instead, each removal produces a meaningful degradation along a specific axis. This pattern is exactly what we would expect from a method whose components are aligned with distinct attack requirements: node selection for budget efficiency, graph-aware generation for robustness under message passing, and feature similarity for stealth.

\subsection{Practical Discussion}
Beyond the numerical comparison, the experimental evidence suggests several practical lessons for the security of TAG systems. First, the threat is not restricted to a single benchmark regime. TAGBD remains effective across graphs that differ substantially in size, text length, and label complexity. This breadth matters because a defense that works only on small citation graphs may not carry over to larger and more heterogeneous applications.

Second, the attack remains viable under limited knowledge. The transfer results show that the attacker does not need exact control over the deployment architecture. This property lowers the operational barrier to attack. In a realistic setting, an adversary may know the task and the type of data being collected, but not whether the final model is a GCN, GraphSAGE, GAT, or a transformer-style graph model. TAGBD still succeeds in that setting because it poisons the training distribution itself rather than overfitting to a single classifier.

Third, the experiments show that stealth should be measured explicitly rather than assumed. It is tempting to regard text poisoning as inherently visible, especially compared with latent-space perturbations. Our results challenge that assumption. When the trigger is generated by a strong text model and injected in a semantics-preserving way, the poisoned content can remain much closer to clean text than naive attack formulations would suggest. This observation is why we report perplexity, text length, and qualitative examples together rather than relying only on ASR.

Finally, the results suggest a useful way to think about attacker strength. Overwriting and appending should not be interpreted simply as two engineering variants of the same method. They correspond to different deployment scenarios. Overwriting approximates cases where the attacker can fully control some training texts, while appending approximates settings where only a small, less intrusive edit is realistic. The fact that appending remains highly successful is particularly important because it shows that even relatively mild-looking edits can be enough to implant a persistent backdoor.




\section{Related Work}

The literature most relevant to our work spans graph backdoor attacks, general adversarial robustness for GNNs, and text-oriented attacks on graph data. Taken together, these studies establish the broader vulnerability of graph learning, but they do not yet provide a complete account of realistic, stealth-oriented text-only backdoor poisoning in TAGs. We briefly review these lines of work to clarify both the intellectual lineage and the novelty of TAGBD.

\textit{Backdoor Attacks on Graph Learning.}
Backdoor attacks implant hidden behaviors during training so that inputs containing a trigger are mapped to an attacker-chosen target label \cite{5bchen2017targeted,4bgu2019badnets,6bliu2020reflection}. In graph learning, the most influential line of work uses structural triggers, for example by injecting fixed or generated subgraphs \cite{sba2021,gta2021,ugba2023,dpgba2024,10.1145/3450569.3463560,272256}. More recent studies also explore stealthier graph triggers and clean-label variants \cite{10.1145/3543507.3583392,GaoLZWJX23clearnLabelBackdoor}. These studies have been essential in showing that graph models can be persistently compromised. At the same time, they largely assume that the attacker can manipulate topology or insert graph patterns that may leave structural traces. This assumption is meaningful in some scenarios, but it is less suitable for TAG applications in which text is easy to edit while links are not. Our work inherits the backdoor perspective from this literature, but shifts the trigger channel from graph structure to raw node text.

A key point of inheritance from this line of work is the idea that the trigger should be reusable and tied to a target label rather than merely causing general prediction degradation. However, once the carrier changes from topology to text, the main technical problem changes as well. Structural triggers are judged largely by graph-level plausibility, whereas text triggers must additionally satisfy linguistic plausibility. This extra requirement is one of the main reasons why a text-only TAG backdoor cannot be viewed as a trivial extension of prior graph attacks.

\textit{Adversarial Robustness of GNNs.}
Another important line of research studies adversarial perturbations to graph models more broadly \cite{oodGL,FeatureBasedGraphBackdoorAttack,gao2020backdoor,11ili2022backdoor,rethinkingbackdoor,GaoLZWJX23clearnLabelBackdoor,GraphAdversarialAttack,dai2024comprehensive}. Early efforts mainly targeted edges, showing that small structural changes can distort message passing \cite{8bDBLP:conf/cikm/TaoCSHWC21,AdversarialAttackGNN2020www}. Later work considered feature-level perturbations and graph backdoors in latent spaces \cite{9bDBLP:conf/nips/LiuLW0L22,ding_spear_2025}. At the defense level, recent studies also examine certified robustness and provable guarantees against backdoors \cite{wang2020certifying,xie2021crfl,weber2023rab}. These contributions provide useful tools and threat models, but most of them operate on inputs that are easier to formalize than to manipulate in real applications. In contrast, our setting asks what happens when the attacker can edit only the natural-language content that the deployed system actually consumes.

This distinction matters when translating robustness analysis into operational security. A perturbation in latent space can be mathematically clean and experimentally informative, yet it may not correspond to an action that a real-world attacker can perform. Text poisoning, in contrast, maps directly to actions such as editing user profiles, posting comments, or submitting metadata. By grounding the attack in raw node text, our work moves the threat model closer to the points where deployed TAG systems actually interact with untrusted inputs.

\textit{Textual Attacks on Graph Data.}
Recent work has begun to investigate text as an attack surface in graph learning \cite{zou2021tdgia,10bDBLP:conf/nips/ZhengZDCYXX0021,lei2024intruding}. Existing studies show that semantic-preserving text perturbations can mislead GNNs and that training-time text corruption can create persistent vulnerabilities. Related work on poisoned or trojaned language models further shows that stealthy text-space attacks can be highly effective even when the perturbation is subtle \cite{kurita2020weight,12ili2021invisible}. This line of research is particularly important because it bridges graph learning with natural-language attacks. However, the current picture is still incomplete: most studies focus on evasion settings, hybrid perturbations, or general poisoning, rather than controllable backdoor injection with explicit trigger design and stealth analysis in TAGs. Our work extends this direction by introducing a graph-aware, text-only backdoor framework that explicitly studies the trade-off between attack effectiveness and linguistic naturalness.

Compared with these lines of work, TAGBD combines three elements in a single framework: explicit targeted backdoor behavior, graph-aware trigger generation, and direct stealth evaluation through both quantitative and qualitative analyses. In this sense, TAGBD lies at the intersection of two inheritances: the graph backdoor literature contributes the notion of persistent trigger-activated malicious behavior, whereas the textual attack literature contributes the emphasis on semantic plausibility and natural-language manipulation. Our main contribution is to connect these two threads in text-attributed graphs and show that stealthy poison text can itself act as a practical and persistent backdoor carrier.

\section{Limitations and Defense Outlook}
TAGBD is evaluated on inductive node classification with a gray-box attacker who can poison a small number of training texts. This setting is realistic and security-relevant, but it does not cover every graph-learning task or every data collection pipeline. In particular, link prediction, graph classification, temporal TAGs, and settings with stricter label control may exhibit different failure modes. We therefore view this paper as establishing a core vulnerability rather than claiming identical behavior across all graph-text workloads.

The results nevertheless suggest a clear defense direction. Future defenses should inspect text and graph structure jointly rather than treat them as separate channels. Promising directions include checking whether node text is semantically consistent with its neighborhood, auditing whether short textual edits exert disproportionate influence on predictions, and building richer stealth metrics beyond perplexity alone. Classical backdoor detection and mitigation ideas such as activation clustering, Neural Cleanse, and anti-backdoor training on poisoned data \cite{chen2018detecting,wang2019neural,li2021anti} may offer useful starting points, but they still need to be adapted to the coupled text-graph setting considered here.

\section{Conclusion}\label{se:conclusions}

This paper presented TAGBD, a graph-aware framework for stealthy poison text backdoors in text-attributed graphs. TAGBD shows that an attacker can implant targeted malicious behavior by poisoning only node texts, without modifying graph topology. By combining uncertainty-guided node selection with the jointly trained TextTrojan generator and shadow GNN, the method produces poison text that is both effective and reasonably natural.
Experiments across datasets, target models, and defense settings show that TAGBD consistently outperforms strong graph backdoor baselines. The two trigger injection modes also reveal a clear design trade-off: overwriting offers maximum attack strength, whereas appending better preserves linguistic naturalness and therefore offers stronger stealth.
Overall, our results demonstrate that inconspicuous text is a realistic and high-impact attack channel in graph learning pipelines. This observation broadens the threat model for TAG systems and suggests that future defenses should reason about graph structure and node content jointly, rather than treating them as separate security problems.

More broadly, the findings of this paper suggest that future TAG security research should move beyond topology-centric threat assumptions. Once text and graph structure are learned jointly, a weakly monitored content channel can be sufficient to implant persistent malicious behavior into the final model. This is particularly important for large, continuously updated systems in which textual content is collected from open environments and only lightly curated. In such settings, even a small amount of stealthy poisoning may be enough to create a reusable trigger-target association. We hope the present study encourages both attackers and defenders to view node content not as auxiliary metadata, but as a first-class security boundary in graph learning systems.


\vspace{-0.3em}
\bibliographystyle{IEEEtran}
\bibliography{bibfile}

\end{document}